# Offline Handwriting Recognition using Genetic Algorithm

Rahul KALA[1], Harsh VAZIRANI[2], Anupam SHUKLA[3] and Ritu TIWARI[4]

[1] Soft Computing and Expert System Laboratory, Indian Institute of Information Technology and Management Gwalior,
Gwalior, Madhya Pradesh-474010, India

[2] Soft Computing and Expert System Laboratory, Indian Institute of Information Technology and Management Gwalior,
Gwalior, Madhya Pradesh-474010, India

[3] Soft Computing and Expert System Laboratory, Indian Institute of Information Technology and Management Gwalior,
Gwalior, Madhya Pradesh-474010, India

[4] Soft Computing and Expert System Laboratory, Indian Institute of Information Technology and Management Gwalior,
Gwalior, Madhya Pradesh-474010, India

**Abstract**
Handwriting Recognition enables a person to scribble something on a piece of paper and then convert it into text. If we look into the practical reality there are enumerable styles in which a character may be written. These styles can be self combined to generate more styles. Even if a small child knows the basic styles a character can be written, he would be able to recognize characters written in styles intermediate between them or formed by their mixture. This motivates the use of Genetic Algorithms for the problem. In order to prove this, we made a pool of images of characters. We converted them to graphs. The graph of every character was intermixed to generate styles intermediate between the styles of parent character. Character recognition involved the matching of the graph generated from the unknown character image with the graphs generated by mixing. Using this method we received an accuracy of 98.44%.

## 1. Introduction

Handwriting recognition refers to the identification of written characters. The problem can be viewed as a classification problem where we need to identify the most appropriate character the given figure matches to. Offline character recognition refers to the recognition technique where the final figure is given to us [Bertolami, Zimmermann and Bunke, 2006]. We have no idea of how the writer wrote the letter. This is contrary to the online character recognition systems where the data can be sampled while the character is being written. An example of this is writing a character on a touch screen with a pointing device. Operating in offline mode gives as input the complete picture of character that we need to recognize. The complexity of the recognition is usually associated with the size of the language being considered. If the language contains more number of characters, the identification would be much more difficult than the case when the language contains lesser number of characters. Similarly we need to consider how the various characters are written and the differences between the various characters. They always have an effect on the performance of the handwriting recognition system.

In this paper we propose the use of Genetic Algorithms for solving this problem. The basic idea of genetic algorithm comes from the fact that it can be used as an excellent means of combining various styles of writing a character and generating new styles. Closely observing the capability of human mind in the recognition of handwriting, we find that humans are able to recognize characters even though they might be seeing that style for the first time. This is possible because of their power to visualize parts of the known styles into the unknown character. In this paper we try to depict the same power into the machines.

In Section 2, we would be discussing the present works and the motivation behind the algorithm. In section 3 we describe the algorithm and its details. The use of Genetic Algorithms is discussed in section 4. In Section 5 we discuss about the testing of the algorithm and its results. Finally, section 6 we give the conclusion.

## 2. Motivation

Handwriting recognition has always been a special problem. The problem increases when we operate it in the offline mode. We see a lot of work has been done in this area in the past few years. The solutions being proposed mainly use Artificial Neural Networks (ANN) and Hidden Markov Models (HMM) for solving the problem. Genetic algorithms have not been applied much. They have been applied for feature selection optimization [Soryani and Rafat, 2006; Shi, Shu and Liu, 1998]. Artificial Neural Networks involve training of the system with all the characters [Draghici, 1997; Yuelong, Jinping and Li, 2006; Som and Saha, 2008; Graves, et. al. 2008]. Then when an unknown input is given to the system, the Artificial Neural Network is able to find out the most probable character by generalization. Hence once trained, the system would be ready to recognize the given unknown input. Hidden Markov Model is a complete statistical model that tries to predict the unknown sequence [Flink and Plotz, 2006; Hewavitharana and Fernando, 2002]. Hence it also tries to recognize the unknown character which is given as input.

For a system to perform well, it is very important to train it well. If the difference between the unknown input and the training data is large, the system may not behave well. Hence there happens a need of giving a diverse training data to the system, depending on what all the system might expect in future. Many recognition systems are author specific, which means that the difference in ways in which the character can be written will not wary much. The training data for a good system hence needs to be designed very well. On the other hand, if we look at ant character from the English language, it can be easily visualized as a graph consisting of lines and curves joined to each other. There are enumerable ways in which a character can be written. E.g. consider writing the letter 'A, in any of the ways given in figure 1. Hence we see that the problem can be very easily visualized as a graph matching problem. Also, as stated above, we also see the fact that various styles of writing can be intermixed to generate new styles. This gives us the motivation to use genetic algorithm to solve the problem of handwriting recognition.

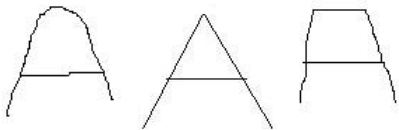

Figure1: Various styles of writing 'A'

## 3. Algorithm

In this section we will take a deep insight into the algorithm and its working. We discuss about the handwriting recognition general procedure, the algorithmic assumptions and its working. We know that we are given an unknown character that needs to be recognized. For this we have diverse form of training data available for each and every character. In this algorithm we try to match the input to the training data and the data generated from intermixing of training data, to find the best match for the given input data.

3.1 General Procedure

Handwriting recognition is a famous problem which involves the recognition of whatever input is given in form of image, scanned paper, etc. The handwriting recognition generally involves the following steps [Liwicki and Bunke, 2007]:

- **Segmentation:** This step deals with the breaking of the lines, words and finally getting all the characters separated. This step involves the identification of the boundaries of the character and separating them for further processing. In this algorithm we assume that this step is already done. Hence the input to our system is a single character.
- **Preprocessing:** This step involves the initial processing of the image, so that it can be used as an input for the recognition system. In this algorithm we assume that a part of this step has been done. We assume that the character segmented is made thin to a unit pixel thickness. Various algorithms may be used for this purpose. The further processing is done by our algorithm.
- **Recognition:** Once the input image is available in good condition, it may be processed for recognition. The role of the recognition system is to identify the character. Our algorithm uses an image as an input for the same.

3.2 Procedure

Once the prerequisites are met, the image input is given to the system. This is then recognized by the algorithm. The algorithm is as given below.

**HandwritingRecognition(Language,TrainingData,InputImage)**
Step1: For every character c in language
Step2:      For every input i for the character c in test data
Step3:           Generate Graph $g_{ci}$ of i
Step4: Generate graph t of input image
Step5: For every character c in language

Step6: Use Genetic Algorithm to generate hybrid graphs
Step7: Return character corresponding to graph with the minimum most fitness function (out of the graphs generated in any genetic operation)

Seeing the previous algorithm it is clear that we first need to generate graphs and then used genetic algorithm to mix these graphs and find the most optimal solution (section 4).

**Generation of Graph:** This algorithm takes as input an image, and returns the graph of the same. The whole procedure of the algorithm requires the principles of graph theory and coordinate geometry. The algorithm is given in figure 2.

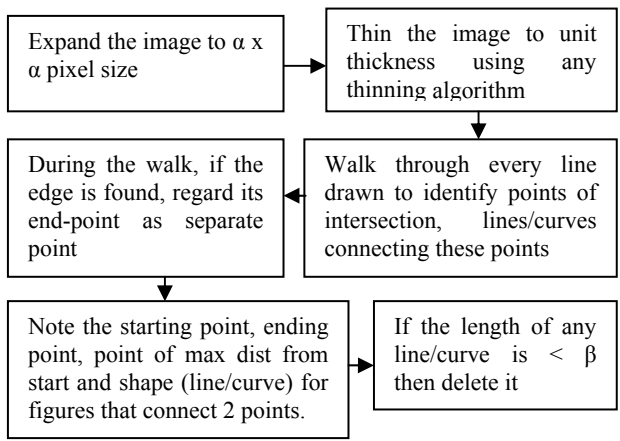

Figure2: The Graph Generation Algorithm

Here a graph represents the vertices and the edges. The edges are the lines or curves connecting any 2 points. Every point where an edge ends/starts is regarded as a vertex. We are also interested in knowing the point for every line/curve, which is at the maximum distance from the start point. This is useful when the graph may contain a closed curve E.g. O would be regarded as a curve from a vertex that ends at the same vertex.

Each edge must hence represent the start vertex, end vertex, shape (line/curve) and the point of maximum separation from start vertex. The image expansion is done by calculating the expansion factor (final image size/initial image size), for both the x and y coordinates. The start and end coordinate of each pixel in the new image are then measured by multiplying with the expansion factor. The lines and curves are differentiated from the maximum and minimum angle subtended by the start of the line/curve, a point situated $\gamma$ units further from the start and all points $2\gamma$ units from the start (Refer Fig.3). For a line the difference between the maximum and minimum angle must be almost 0 degrees. Law of cosines is used for the purpose of finding angles. Edges are detected by using a similar logic. Figure 4 shows the graph generated when the input was J.

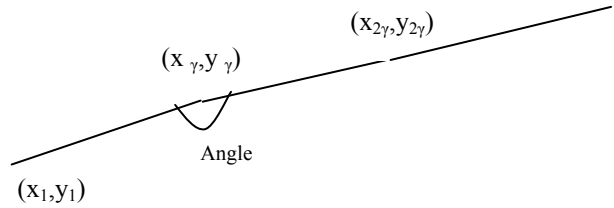

Figure3: Difference between curve and line

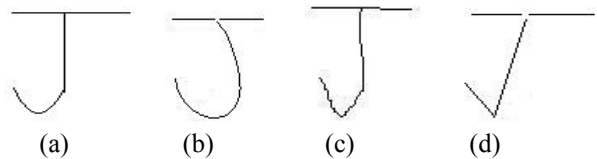

Figure4: Two Ways J was written (a, c) and the generated graph (b, d)

## 4. Genetic Algorithm

Genetic algorithms are a very good means of optimizations in such problems. They optimize the desired property by generating hybrid solutions from the presently existing solutions. These hybrid solutions are added to the solution pool and may be used to generate more hybrids. These solutions may be better than the solutions already generated. All this is done by the genetic operators, which are defined and applied over the problem. We already have a set of graphs generated from training data for any character. The use of genetic algorithm is to mix 2 such graphs and to generate new graphs. These newly generated graphs may happen to match the character better than the existing graphs. Hence genetic algorithms are a good means of optimizations. We discuss each of the points in detail in the coming sections.

4.1 Fitness Function

In Genetic Algorithms, the fitness function is used to test the goodness of the solution. This function, when applied on any of the solution from the solution pool, tells the level of goodness. In our problem, we have used fitness function to measure the deviation of the graph of the solution, to that of the unknown input. If the two graphs are very similar, the deviation would be low and hence the value of the fitness function would be low. The lower the value of the fitness function, the better would be the matching. Hence the graph with the lowest value of fitness function would be the most probable answer. We first devise a formula to find the deviation between any two edges. This would be then used as a means of finding the deviation between two graphs.

***4.1.1 Deviation between two edges:*** For finding out the deviation between two edges, we first define a function $D(e_1,e_2)$ that finds the deviation between any edges of a graph ($e_1$) with any edge of the other graph ($e_2$). Here an edge may represent a line or a curve. But the start point of the edge $e_1$ may match with the start point of the edge of $e_2$ and the end point of $e_1$ may match with the end point of $e_2$. It may also be possible for the converse to be true. The start point of $e_1$ may match with the end point of $e_2$ and the end point of $e_1$ may match with the start point of $e_2$. This is shown in figure 5(a)-(c). Hence we calculate the deviation using two separate cases ($D_1$ and $D_2$) and the minimum of the two is the actual deviation. $D_1$ represents the case where the start vertex of $e_1$ matches with the start vertex of $e_2$. The end vertex of $e_1$ matched with the end vertex of $e_2$. In general

$D_1(e_1,e_2)$=square of distance of start points of $e_1$ and $e_2$ + square of distance of end points of $e_1$ and $e_2$

If however, $e_1$ is a line and $e_2$ is a curve or vice versa, an overhead cost of $\eta$ is added. If both $e_1$ and $e_2$ are curves, and start and end points of $e_1$ are less than $\beta$ units apart (it is almost circle), then we take point of maximum distance in place of end points in the above formula. Here point of maximum distance is the point in the curve which is at maximum distance from the start point of the curve. This is shown in Figure 5(a) and Figure 5(b).

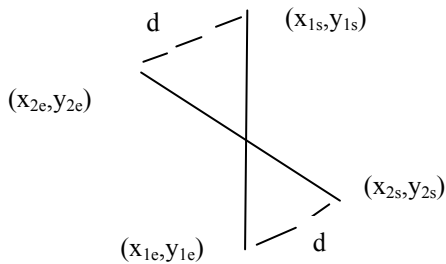

Figure 5(a): Calculating D1 with 2 lines

$D_2$ represents the case where the start vertex of $e_1$ matches with the end vertex of $e_2$. The end vertex of $e_1$ matched with the start vertex of $e_2$. Similarly we calculate $D_2(e_1,e_2)$ by the following formula

$D_2(e_1,e_2)$=square of distance of start point of $e_1$ and end point of $e_2$ + square of distance of end point of $e_1$ and start point of $e_2$

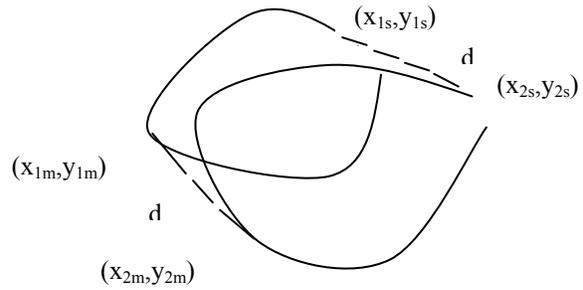

Figure 5(b): Calculating D1 with 2 curves

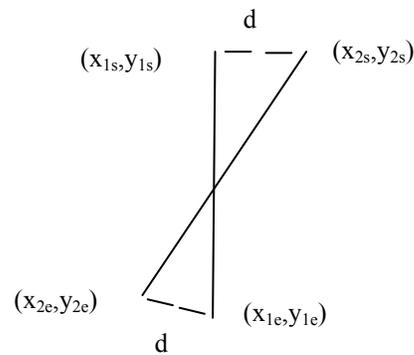

Figure 5(c): Calculating D2

Other specifications remain same as used in calculating $D_1(e_1,e_2)$. This is shown in Figure 5(c). The deviation between two edges is calculated by the following formula:

$D(e_1,e_2)=\min\{D_1(e_1,e_2),D_2(e_1,e_2)\}$

We even generalize the formula to the condition when either $e_1$ or $e_2$ is null. This means that we can find the deviation of a line or a curve with nothing. This is a feature useful in finding graph deviation when there is unequal number of edges in two graphs. In such cases the formula is:

$D(e_1,null)$ or $D(null,e_1)$ = Distance between the starting point and end point of line/curve.

If $e_1$ is curve, and start and end points of $e_1$ are less than $\beta$ units apart (it is almost circle), then we take point of maximum distance in place of end points in the above formula. Here point of maximum distance is the point in the curve which is at maximum distance from the start point of the curve. Suppose that the edge $e_1$ in first graph has start points as $(x_{1s},y_{1s})$ and end points as $(x_{1e},y_{1e})$.

Similarly suppose that the edge $e_2$ in second graph has start points as $(x_{2s},y_{2s})$ and end points as $(x_{2e},y_{2e})$. The point of maximum distance of $e_1$ is $(x_{1m},y_{1m})$ and $e_2$ is $(x_{2m},y_{2m})$. The formula explained above can also be stated as:

$D_1(e_1,e_2) = (x_{1s}-x_{2s})^2+(y_{1s}-y_{2s})^2+(x_{1e}-x_{2e})^2+(y_{1e}-y_{2e})^2$  if ($e_1$ is line and $e_2$ is line) or ($e_1$ is curve and $e_2$ is curve and the distance between $(x_{1s},y_{1s})$ and $(x_{1e},y_{1e})$ is less than β units)

$D1(e_1,e_2) = (x_{1s}-x_{2s})^2+(y_{1s}-y_{2s})^2+(x_{1e}-x_{2e})^2+(y_{1e}-y_{2e})^2 + \eta$ if ($e_1$ is line and $e_2$ is curve) or ($e_1$ is curve and $e_2$ is line)

$D_1(e_1,e_2) = (x_{1s}-x_{2s})^2+(y_{1s}-y_{2s})^2+(x_{1m}-x_{2m})^2+(y_{1m}-y_{2m})^2$  if ($e_1$ is curve and $e_2$ is curve and the distance between $(x_{1s},y_{1s})$ and $(x_{1e},y_{1e})$ is less than β units)

$D_2(e_1,e_2) = (x_{1s}-x_{2e})^2+(y_{1s}-y_{2e})^2+(x_{1e}-x_{2s})^2+(y_{1e}-y_{2s})^2$  if ($e_1$ is line and $e_2$ is line) or ($e_1$ is curve and $e_2$ is curve and the distance between $(x_{1s},y_{1s})$ and $(x_{1e},y_{1e})$ is more than β units)

$D_2(e_1,e_2) = (x_{1s}-x_{2e})^2+(y_{1s}-y_{2e})^2+(x_{1e}-x_{2s})^2+(y_{1e}-y_{2s})^2 + \eta$ if ($e_1$ is line and $e_2$ is curve) or ($e_1$ is curve and $e_2$ is line)

$D_2(e_1,e_2) = (x_{1s}-x_{2m})^2+(y_{1s}-y_{2m})^2+(x_{1m}-x_{2s})^2+(y_{1m}-y_{2s})^2$  if ($e_1$ is curve and $e_2$ is curve and the distance between $(x_{1s},y_{1s})$ and $(x_{1e},y_{1e})$ is less than β units)

$D(e_1,e_2)=\min\{D_1(e_1,e_2),D_2(e_1,e_2)\}$ if both $e_1$ and $e_2$ are not null

$D(e_1,e_2)= (x_{1s}-x_{1e})^2+(y_{1s}-y_{1e})^2$  if $e_2$ is null

$D(e_1,e_2)= (x_{2s}-x_{2e})^2+(y_{2s}-y_{2e})^2$  if $e_1$ is null

*4.1.2 Deviation of a graph:* Once we know the deviation of an edge with another edge, the deviation of a graph can be found out easily. This deviation is found out by pairing up of edges and iterating through all the edges. The algorithm is as given below.

**Deviation($G_1$,$G_2$)**
Step1: dev ← 0
Step2: While $G_1$ has no edges or $G_2$ has no edges
Step3:     Find the edges $e_1$ from first graph and $e_2$ from second graph such that deviation between $e_1$ and $e_2$ is the minimum for any pair of $e_1$ and $e_2$
Step4:     Add its deviation to dev
Step5:     Remove $e_1$ from first graph and $e_2$ from second graph
Step6: For all edges $e_1$ left in first graph
Step7:     Add deviation of $e_1$ and null to dev
Step8: For all edges $e_2$ left in second graph
Step9:     Add deviation of null and $e_2$ to dev
Step10: Return dev

Here we see that we try to minimize the total deviation. For this we select the pair of edges, one from each graph such that their deviation is minimal. We keep selecting such pairs till one graph gets empty. Hence we keep proceeding by keeping the total deviation minimum. In the end we add deviation of all the left edges or curves. This way we find the minimum deviation between the two graphs.

4.2 Crossover

Crossover is the operation in genetic algorithms where we mix two solutions and form a new solution. This may be better than the two existing solutions. The crossover operation helps us to generate newer solutions and thus helps in optimization.

In this problem we have a graphical representation of the solution sets. Each solution is a graph which contains both edges and vertices. The crossover operation uses two graphs to mix them and forms a new graph. This graph takes some characteristics from the first graph and some characteristics from the second graph. The graph generated from this operation may be better than the parent graph and hence useful. The basic motive of using this operation is to mix styles. If the two graphs have characters in different styles, we would be able to mix them and form a style that is intermediate between the parent styles. The crossover operation makes sure that the style of writing a particular section of the character is taken from one of the graph. This section is removed from the other graph and the new section is added. Hence using the crossover operation we may be able to mix styles to form unique new solutions. Many solutions are possible for every combination of parents. In this algorithm, we generate all the forms and add it as a solution. Hence one crossover operation results in many solutions being generated.

The crossover operation in any pair of graphs is always carried out between two vertices of the graph. Hence we need to find out the pair of points using which the crossover needs to be done. We use the first graph as the base. We try to find paths that connect these two points. We lay down a condition that any point can be visited at most once while finding the path. Also the total path length must be less than 4 edges. This means we must be able to reach the second vertex from the first, using a maximum of 3 intermediate distinct vertices. Once such a path is found in the first graph, we do a similar work using the same points in the second graph as well. If the path is found, we are ready for crossover. The final generated solution is same as the first graph. The only difference is that we delete the entire path that was found between the two chosen points from first graph. We then insert the path that was chosen from the second graph between the same chosen points. Hence a region of style between the chosen points is changed from the old style of first graph to the new style of graph two. The style of the remaining character remains same as graph one. However a point in the first graph may not be the same in the other graph as well. For this there happens to be a need for matching the points. A point of first graph is tried to match to the closest possible point in the second graph. The points that fail to match to any other point are added in the end to the other graph as well. This means that if a point in graph one failed to match any other point in graph two, it would be simply added as a vertex in graph two. The following is the algorithm used for the crossover of the two graphs

**CrossOver($G_1$,$G_2$)**
Step1: match ← MatchPoints($G_1$,$G_2$)  (Find points in $G_1$ corresponding to $G_2$ and vice versa)
Step2: $W_1$ ← GenerateAdjacency($G_1$)  (Generates the Adjacency Matrix of the Graph)
Step3: $W_1$ ← GenerateAdjacency($G_2$)
Step4: if $W_1 \neq W_2$
Step5: FindPaths($W_1$)  (Finds the paths of all lengths between any pair of points)
Step6: FindPaths($W_2$)
Step7: $p_1$ ← path between vertices $v_1$ and $v_2$ in $G_1$ of length l, for all $v_1$, $v_2$ in $W_1$ and length l= 1 or 2 or 3 or 4
Step8: $p_2$ ← path between vertices match[$v_1$] and match[$v_2$] in $G_2$ of length $l_2$=1 or 2 or 3 or 4
Step9: If $p_1$ exists and $p_2$ exists and $p_1 \neq p_2$
Step10: Remove all edges from $W_1$ that are found in $p_1$
Step11: Add all edges in $W_1$ that are found in $p_2$
Step12: g ← MakeGraph($W_1$,$W_2$) (Generate graph out of Adjacency matrix of $W_1$)
Step13: Add g to solution set
Step14: else g ← MakeGraph($W_1$,$W_2$)
Step13: Add g to solution set

Each of the steps is discussed below in the further sections

*4.2.1 Matching of points:* First we define the algorithm for the matching of points in the two graphs. If a point in graph 1 is said to match a point in graph 2, it may be assumed that a reference to the point in graph 1 would be taken analogous to the reference of the matching point in graph 2. Suppose both graph 1 and graph 2 are the character I written in two different styles. Suppose the point $s_1$ in graph 1 is the leftmost point of the topmost line. If we say that this point $s_1$ matches with the point $s_2$ in graph 2, then we may assume the point $s_2$ in graph 2 to be the leftmost point of the topmost line. This is shown in figure 6. The algorithm is as given below.

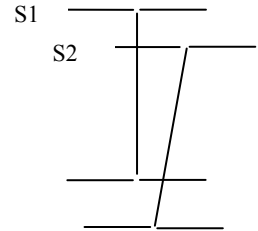

Figure 6: The matching of points in 2 graphs

**MatchPoints($G_1$,$G_2$)**
Step1: $ver_1$← all unique points in first graph which are at least β units distance apart from each other
Step2: $ver_2$← all unique points in second graph which are at least β units distance apart from each other
Step3: match ← null
Step3: While $ver_1$ is not null or $ver_2$ is not null
Step4: match[$s_2$] ← $s_1$ such that $s_1$ is a vertex in $ver_1$ and $s_2$ is a vertex in $ver_2$ and distance between $s_1$ and $s_2$ is least for any combination of vertices in $ver_1$ and $ver_2$ and distance between $s_1$ and $s_2$ is less than 2β units
Step5: Remove $s_1$ from $ver_1$
Step6: Remove $s_2$ from $ver_2$
Step7: For all vertex v in ver1
Step8: match[v] ← v
Step9: For all vertex v in $ver_2$
Step10: match[v] ← v

*4.2.2 Generate Adjacency Matrix:* Once the points have been identified and matched, the next step is to generate the graph. We use the adjacency matrix way of graph representation for this problem. We know that since this is a non-directional graph, the path from a node x to a node y would also imply the path from the node y to the node x. This means the final adjacency matrix (W) would be symmetric. Further, we know that any two points can be simultaneously connected by a curve and/or a line. Hence we use the following convention for the representation of a cell $W_{ij}$ in the adjacency matrix for any pair of vertices i and j.

$W_{ij}$ = 0 (If the node i is not directly connected by the node j)
$W_{ij}$ = 1 (If the node i is directly connected by the node j by only 1 line)
$W_{ij}$ = 2 (If the node i is directly connected by the node j by only 1 curve)

$W_{ij}$ = 3 (If the node i is directly connected by the node j by only 1 line and 1 curve)

All other states are invalid. The algorithm for the generation of the adjacency matrix is given below.

**Generate Adjacency(G)**
Step1: W ←0
Step2: For all edges e in graph joining points i and j
Step3: $W_{ij}$ ← $W_{ij}$+1 (if e is a line)
Step4: $W_{ij}$ ← $W_{ij}$+2 (if e is a curve)
Step5: $W_{ji}$ ← $W_{ij}$
Step6: Return W

*4.2.3 Find Paths:* Once we have generated the adjacency matrix, the next step is to find the paths of all lengths between all pairs of points. This would be required in the working of the algorithm. The data generated by this step becomes a straight input for the further step. We use the concept of dynamic programming to solve this problem.

We know that we have a graph. This consists of a number of vertices. For this problem we restrict ourselves to paths which are of a maximum length of 4. Hence the problem is to find the existence of a path p, from a vertex i to a vertex j of a distance of exactly l. Here i and j can take any values and l is 1, 2, 3 or 4. We define a variable $F(i,j,l)$ that stores the existence of the path. It stores the last vertex traversed while travelling from i to j using path of length l. If such a path is not possible, it stores null

We also know that for single length (i=1), path is only possible if there is a direct edge between the vertex i and vertex j. Hence

$F(i,j,l)$ is defined as follows

$F(i,j,0)$=i, if edge (i,j) exists or edge (j,i) exists
$F(i,j,0)$=null, otherwise
And
$F(i,j,l)$= null, if there is no path possible of length l (l ≠ 0) between vertex i and vertex j.
$F(i,j,l)$= k, if there is a path possible of length (l-1) between vertex i and vertex k and j is directly connected to j(l≠0)

This can also be written as
$F(i,j,l)$= k, if $F(i,k,l-1)$ exists and (vertex (k,j) exists or vertex (j,k) exists), l ≠ 0
$F(i,j,l)$= null, otherwise

Hence using this relation we can find whether the path of length l exists between nodes i and j. If it exists, the value of $F(i,j,l)$ is not null. If the path exists, we can find the path by reverse iterating the value of $F(i,j,t)$ by varying t from l to 0. This way we would be able to find the exact path that joins the two vertices.

However while executing the algorithm we need to take care of the following points:

- Any vertex can occur maximum once in any path. Hence before making any selection of k in the above formula of $F(i,j,l)$, we need to make sure that we do not include a point twice in the entire path from i to j.
- It is possible for various values of k to satisfy the above formula of $F(i,j,l)$. In such a situation, we first find out all such possible values of k and then select any one randomly.

*4.2.4 Removing and adding edges:* As mentioned in the above algorithm of crossover, we apply the above data to find paths in first and second graph between any two points. If both the paths are found, we remove the edges of first path and add the edges of the second path from $G_1$. In such a manner we are able to mix two distinct graphs and generate a new graph.

However there may be various possibilities in the addition or removal. While removing an edge from vertex i to a vertex j in the graph, we take care of the following conditions, to maintain the consistency of the graph

If $W_{ij}$=0, or in other words there exists no edge from i to j, we break the operation
If $W_{ij}$=1 or in other words there exists a line from i to j, we remove the line
If $W_{ij}$=2 or in other words there exists a curve from i to j, we remove the curve
If $W_{ij}$=3 or in other words there exists a curve and a line from i to j, we remove any one of them randomly choosing
$W_{ij}$>3 is an illegal state and the operation is broken

After the operation $W_{ij}$ is made equal to $W_{ji}$

Here $W_{ij}$ refers to the adjacency matrix of graph 1 on which the removal operation is applied. Here we have taken care that after the operation is over, $W_{ij}$ should be between 0 and 3, which are the only legal values it can take.

Similarly while adding an edge from vertex i to a vertex j in the graph, we take care that after the operation, $W_{ij}$ should be between 0 and 3, and $W_{ij}$ should be equal to $W_{ji}$.

Intermixing of graphs may generate a lot many impossible graphs. By proper checking and breaking operations, we save the solution set from getting wrong data.

*4.2.5 Generation of graph:* The last step is to convert the graphs in form of Adjacency Matrix into graphs in form of edges list. Here we need to take care of another kind of optimization. This is the distance optimization of the algorithm. As explained above, in the algorithm we found out the matching vertices between the two graphs. In this method, we matched a vertex of first graph to the closest one of the second graph and vice versa. The purpose of this matching was to say that matching vertices are analogous to each other across the two graphs. Hence we said that if a vertex $s_1$ in first graph matches with a vertex $s_2$ in second graph, this means that the role of $s_1$ in first graph is played by $s_2$ in second graph.

But we need to realize that the fitness function works on the separation between the vertices. While reforming the graph in this step, we would be free to take the vertices of first graph as the final vertices of the mixed graph. We may also take the matching vertices of the second graph in place of those of the first graph. But since we have already calculated the fitness function value using the parents, using any of these methods only gives us the benefit of style optimizations. If the vertices of the unknown input lie in between those of the parents, these methods would not serve very useful.

In pace of taking vertices from any one of the graph, we take the mean of the matching vertices of the two graphs. This optimizes the distance as well, along with the style. It may be noted that in case the two graphs match in style, which is the case most number of times, the distance optimization proves very useful in optimizations. The algorithm for the formation of graph is given below

**GenerateGraph(W)**
Step1: for all vertex i W
Step2:              for all vertex j in W that come after or at i
Step3:                    if $W_{ij} > 2$
Step4:                          add a curve from x to y
Step5:                          $W_{ij} \leftarrow W_{ij} - 2$
Step6:                    if $W_{ij} = 1$
Step7:                          add a line from x to y

Here x=mean of vertices i in first graph and the corresponding matching vertex of i in second graph (vertex k such that match[k] ← i)

y=mean of vertices j in first graph and the corresponding matching vertex of j in second graph (vertex k such that match[k] ← i)

## 5. Results

In order to check the working of the algorithm, we coded the algorithm. Java was used as a language. Input was given in the form of images. Test data was also stored in the form of images. We applied the algorithm over the capital letters of English language. The language contains 26 characters. In order to make the database, we first wrote each letter twice. The first time the letter was written perfectly, like the way we are taught. The second time we used a raw hand to introduce some imperfection. Letters like A had some types of style associated with them. We hence, wrote these letters more times as per requirements. The input data set was made by writing each character enumerable number of times, using different ways. The motion of the hand was shaken while writing, to introduce some unknown imperfections in the test data. The algorithm was made to first process the training data. Then it was made to run on the test data by iterating through the test data one after the other. We used a total of 69 characters as the training data. The test data was a collection of 385 inputs. When the algorithm was made to run, it correctly identified 379 characters. It wrongly identified 6 test cases. This gave the efficiency of 98.44%.

This problem could have been solved with the absence of Genetic Algorithms as well. In order to see the importance of Genetic Algorithms in the problem, we tested the data in the absence of the application of genetic algorithms. We found that the genetic algorithms were useful in the following manner

### 5.1 Distance Optimization

The character that is input for identification may have the vertices of its graph at a point quite far from that of the training data. Actually language allows us to end the various lines over a large distance. This depends on the writer and his writing style. But when we would compare the closeness of such an input, it is clear that the differences would be large. On the other hand identification by a human of the same character would be very easy. Such a problem was highlighted when we tried to recognize the character M without Genetic Algorithm. The input given is shown in Figure 7(a). As it can be seen in this figure, the two slanting lines of M are quite medium in size and little near. This was not true in any of the training data. The characteristic of the training data was that one contained the entire M distributed normally. This made the slanting lines quite big. Another one was written in a quick jerk to the middle section, making the section quite short. This is shown in figure 7(b) and (c). When the algorithm was made to run, it happened to match with the character X. But on applying genetic algorithms, the distance got optimized, and the character was identified correctly.

5.2 Style Optimization

We have already discussed that the algorithm is very efficient in mixing two styles and generating a mixture of styles. This property of the algorithm to generate newer and newer styles proves very useful in the working of the algorithm. In the training data set, we had given two distinct Bs in the training data of B. When operated without genetic algorithms, these two failed to match with the input of the data. On the contrary the best matching was found out to be with the letter S with a high deviation. But when the same was done with genetic Algorithms, the B matched correctly. Figure 8 shows the graphs of input, the training data and one of the genetically produced results. Here the top curve of first training data was added to the second training data's top set of lines. The genetic algorithm was applied across the vertical top line.

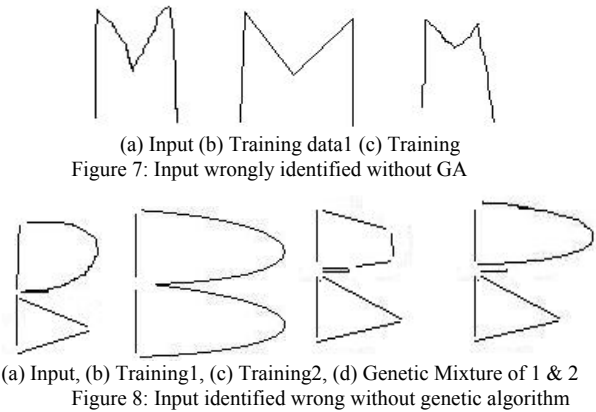

(a) Input (b) Training data1 (c) Training
Figure 7: Input wrongly identified without GA

(a) Input, (b) Training1, (c) Training2, (d) Genetic Mixture of 1 & 2
Figure 8: Input identified wrong without genetic algorithm

## 6. Conclusions

In this paper we proposed the use of genetic algorithm and graph theory for solving the problem of offline handwriting recognition. We had given the input in the form of images. The algorithm was trained on the training data that was initially present in the database. The training data consisted of at least two training data sets per character in the language. We used the graph theory and coordinate geometry to convert the images to graphs. We saw that these conversions changed the whole problem of handwriting recognition to the problem of graph matching. When a pure graph matching was done, sufficiently fine results were obtained. The algorithm could recognize unknown characters given as input. But the efficiency improved drastically when we applied genetic algorithms. This algorithm helped in both style optimization and distance optimization. In style optimization, it helped us to mix two different styles to generate a new one that was in between the two. This was done by mixing two graphs across two points. We saw how it helped in identification of character B. We also saw how the algorithm helped in distance optimization. It transformed the start and end point of vertices in such a way that it could match better with the unknown data input. This was done by taking mean coordinates of the vertices of parents. We saw how it helped in identification of letter M. In all we got an efficiency of 98.44%, which proves that this algorithm works for most of the cases and correctly matched the unknown input to their character.

**Acknowledgments**

This work has been supported and sponsored by Indian Institute of Information Technology and Management Gwalior.

**R. Kala** Mr. Rahul Kala is a student of final year of 5-year Integrated Post Graduate Course (BTech + MTech in IT) in Indian Institute of Information Technology and Management Gwalior. His areas of research are hybrid system design, robotic planning, design of algorithms, artificial intelligence and soft computing. He has published over 25 papers and 2 books in various international and national journals/conferences. He also takes keen interest in Free/Open Source Software.

**H. Vazirani** Mr. Harsh Vazirani is a student of final year of 5-year Integrated Post Graduate Course (BTech + MTech in IT) in Indian Institute of Information Technology and Management Gwalior. His areas of research are artificial neural networks, hybrid system design, speaker recognition, artificial intelligence and soft computing.

**A. Shukla.** Dr. Anupam Shukla is an Associate Professor in the ICT Department of the Indian Institute of Information Technology and Management Gwalior. He completed his PhD degree from NIT Raipur, India in 2002. He did his post graduation from Jadavpur University, India. He has 22 years of teaching experience. His research interest includes Speech processing, Artificial Intelligence, Soft Computing and Bioinformatics. He has published around 120 papers in various national and international journals/conferences. He is referee for 4 international journals and in the Editorial board of International Journal of AI and Soft Computing. He received Young Scientist Award from Madhya Pradesh Government and Gold Medal from Jadavpur University.

**R. Tiwari** Dr. Ritu Tiwari is an Assistant Professor in the IT Department of Indian Institute of Information Technology and Management Gwalior. Her field of research includes Biometrics, Artificial Neural Networks, Speech Signal Processing, Robotics and Soft Computing. She has published around 60 papers in various national and international journals/conferences. She has received Young Scientist Award from Chhattisgarh Council of Science & Technology and also received Gold Medal in her post graduation.